# Bridging the gap in online hate speech detection: a comparative analysis of BERT and traditional models for homophobic content identification on X/Twitter

**Josh McGiff[1] and Nikola S. Nikolov**

Department of Computer Science and Information Systems, University of Limerick, Ireland

[1]Josh.McGiff@ul.ie

**Abstract.** Our study addresses a significant gap in online hate speech detection research by focusing on homophobia, an area often neglected in sentiment analysis research. Utilising advanced sentiment analysis models, particularly BERT, and traditional machine learning methods, we developed a nuanced approach to identify homophobic content on X/Twitter. This research is pivotal due to the persistent underrepresentation of homophobia in detection models. Our findings reveal that while BERT outperforms traditional methods, the choice of validation technique can impact model performance. This underscores the importance of contextual understanding in detecting nuanced hate speech. By releasing the largest open-source labelled English dataset for homophobia detection known to us, an analysis of various models' performance and our strongest BERT-based model, we aim to enhance online safety and inclusivity. Future work will extend to broader LGBTQIA+ hate speech detection, addressing the challenges of sourcing diverse datasets. Through this endeavour, we contribute to the larger effort against online hate, advocating for a more inclusive digital landscape. Our study not only offers insights into the effective detection of homophobic content by improving on previous research results, but it also lays groundwork for future advancements in hate speech analysis.

## 1. Introduction

Hate speech is a pervasive issue on social media, offering a platform for rapid information dissemination but also for harmful behaviours. This environment fosters aggression, discrimination, and a hostile atmosphere, particularly affecting marginalised groups and undermining equality and freedom of expression. Among various types, homophobia stands out, stemming from global disparities in LGBTQIA+ rights, influenced by cultural, educational, and legislative factors [1].

Our research adopts sentiment analysis to detect homophobic hate speech on social platforms, addressing a significant gap in current research that often overlooks this area. Sentiment analysis, a key technique in natural language processing, is employed to identify negative attitudes and emotions, aiming to classify and mitigate homophobic content effectively. Despite existing efforts to combat other forms of hate speech, such as sexism [2] or racism [3], homophobia remains underrepresented in detection models and general hate speech research. The Google Scholar query 'sentiment analysis machine learning' followed separately by 'racism', 'sexism' and 'homophobia' returns vastly different results of 139k, 37k and 23k respectively. Our research aims to bridge this gap by developing and implementing advanced sentiment analysis models specifically tailored for identifying homophobic textual content.

This endeavour not only aims to enhance hate speech detection and classification of homophobia, but also to contribute to the broader efforts against online hate, promoting a safer, more inclusive digital environment. We hope that the combination of publishing our results for a variety of models, a readily available BERT-powered (Bidirectional Encoder Representations from Transformers) model that outperforms existing research and the largest (to our knowledge) open-source labelled English dataset for homophobia detection, can help to combat homophobia-charged discrimination online.

## 2. Related Work

We explored similar approaches for hate speech detection of different domains such as sexism and racism. One of the fundamental steps in this homophobia detection experiment is to determine the definition of homophobia to be employed by the annotators of our dataset, as it consequently impacts the nature and effectiveness of the classification model. Homophobia can be defined as a bias against homosexuals or behaviour that discriminates against them [4]. Glick and Fiske's ambivalent sexism theory [5] has been used for multiclass classification whereby the theory is applied to sexism detection [2]. This theory postulates that sexism can be divided into hostile and benevolent sexism. Their paper indicates that hostile sexism denotes a characterisation of unfiltered and explicitly negative attitude. Whereas benevolent sexism aligns itself with a more subtle style of displaying prejudice or discrimination. We find that applying this theory to gather both benevolent and hostile homophobic tweets under a common label would yield a suitable definition for qualifying hate speech.

Among the limited research on homophobia detection is a paper that produces a multilingual dataset and experiments for detecting homophobia and transphobia in YouTube comments [6]. While this paper meticulously documents their methods in using a variety of features such as BERT embeddings to train a variety of classifiers, it lacks representative data for homophobia in English. Common hateful terms that are indicative of the existence of homophobia [7] are not included in their dataset. Although their multiclass approach includes transphobia detection, they only collect a handful of transphobic comments in English. This reflects our own challenges in acquiring an adequate dataset of transphobic tweets, leading us to concentrate solely on homophobia.

Preprocessing is critical, as demonstrated by Kwok and Wang's experiment with Naïve Bayes [3] and Jha and Mamidi's use of multiple classifiers such as Support Vector Machine (SVM) and FastText [2]. They describe the data cleaning process involving the removal of URLs, mentions, stopwords, and punctuation. As for the models typically used in hate speech detection, SVM and Long Short-Term Memory (LSTM) models have been contrasted in Facebook hate speech detection [8], noting LSTM's superiority in capturing nuanced context. Similarly, in the context of sentiment analysis, the efficacy of Recurrent Neural Networks over SVM has been demonstrated for analysing Yelp and IMDB reviews [9]. Moreover, the study explores the application of advanced models like BERT and FastText in sentiment classification. Given the short nature of tweets and that BERT has been found to be effective in sentence-level tasks [10] [11], we attempt to exploit BERT's effectiveness at sentence-level classification for our homophobia detection model. Overall, these varying approaches highlight the evolving landscape of sentiment analysis, balancing traditional machine learning with emerging deep learning techniques.

## 3. Dataset

A pivotal aspect of this paper relates to the collection, annotation and publication of the dataset that acts as the input data for training our machine learning models. Unlike widely researched domains such as racism and sexism detection, we were only able to source a handful of publicly available and open-source hate speech datasets that specifically relate to homophobia when we began this research. While many social media platforms such as Facebook and YouTube would be suitable sources of data for this task, X (i.e X.com, formerly known as Twitter) previously provided APIs for extracting data from their platform. This feature combined with X being a medium for users to express open ended standalone content renders it to be an appropriate source to scrape data. Therefore, our approach involves scraping X by enlisting the command line tool Twarc for extracting data from the platform. Given our domain, we refined our scraping technique by filtering for terms and accounts associated with the LGBTQIA+ community.

Our annotation process involved enlisting three volunteers to label the collated dataset. We were fortunate to find volunteers consisting of a range of sexualities and gender identities. Given that participants were labelling content as homophobic or not, we hypothesised that it would be highly insightful to observe the variation in annotations depending on the identities of the participants themselves and to consequently reduce possible biases associated with such identities. A majority vote was used to select the label assigned to each tweet. Due to the potential for disturbing and upsetting content, each volunteer was permitted to cease participation at any time. Participants used Microsoft Excel to label the dataset in batches over several days.

## 4. Experiments

In our experiments, we investigate the impact of BERT embeddings as a feature for classifying homophobic content with traditional models and BERT itself. The BERT approach represents a significant departure from traditional text processing techniques such as Term Frequency-Inverse Document Frequency (TF-IDF) due to its tokenization process. BERT's process of breaking down documents into words and subwords enables the model to build a deeper understanding of language, capturing intricacies that older techniques miss. One of its advantages is that it can capture the context surrounding terms in sentences [12]. This feature of the model is attributed to its bidirectional self-attention mechanism, thus allowing for nuanced dependencies to be tracked by considering both immediate surroundings and wider sentence context. In our study, we employed several machine learning techniques trained on BERT embeddings for our homophobia detection text classification task:

- Logistic Regression (LR): We enlisted LR for tweet classification in order to take advantage of its simplicity and efficiency in binary classification problems such as homophobia detection. This process was executed using Python's sklearn library, which facilitated the application of logistic regression to the high-dimensional data provided by BERT embeddings, maintaining a balance between model complexity and interpretability.

- Naïve Bayes (NB): Despite its simple nature and strong independence assumptions, we applied NB for its efficiency in categorising tweets. Implemented via sklearn, NB served as a comparative baseline for text classification effectiveness.

- Decision Tree (DT): We utilised DT for its ability to classify data based on feature importance. Our implementation, facilitated by sklearn, explored the impact of data variations on classification.

- Random Forest (RF): As an ensemble method, RF combines multiple decision trees to improve classification accuracy through majority voting. Chosen for its robustness and effectiveness, our RF model was also implemented using sklearn.

- Support Vector Machines (SVM): For tweet classification, we utilised SVM, leveraging the kernel trick for optimal data separation. The BERT embeddings were fed into the SVM model to distinguish between homophobic and non-homophobic content, utilising Python's sklearn library.

- TFBertForSequenceClassification: We utilised this BERT variant, which adds a sequence classification head on top of BERT itself that enables prediction with contextual embeddings [13]. Our data preparation involves text tokenization, special token addition ([CLS], [SEP]), and padding/truncation to align with BERT's 768-dimensional token embeddings.

We experimented with both k-fold cross-validation with k=8 and repeated holdout for each of the classifiers as a means to investigate the impact of the evaluation methods on our results. Our holdout method involves averaging the metrics of five different attempts at training a model for a 70:30 train-test split. Although our relatively small dataset of tweets could suggest that cross-validation would be better, we were interested in determining if its balanced nature would allow the holdout method to excel.

## 5. Results and Discussion

In this section, we present and evaluate the performance of a variety of machine learning models, both traditional and BERT-based, trained on BERT embeddings for the binary classification task of identifying homophobic tweets. We chose to focus on the performance of BERT embeddings due to their known effectiveness in capturing the context of words by considering surrounding text. We employed and compared two validation methods to assess the performance of each model: k-fold cross validation and holdout. After experimenting with a range of values of k, specifically 5, 8, 10 and 12, for k-fold cross-validation, we found that k=8 yields the strongest results. On the other hand, we found that a train-test split of 70:30 produces the best results with the holdout approach. We report the accuracy, precision, recall and F1-scores for this classification task in Table 1 and Table 2.

**Table 1.** Model performance with stratified cross-validation (k=8).

| Model | Accuracy | Precision | Recall | F1-score |
|---|---|---|---|---|
| **LR** | 0.768 | 0.776 | 0.739 | 0.755 |
| **NB** | 0.894 | 0.483 | 0.887 | 0.624 |
| **DT** | 0.656 | 0.642 | 0.660 | 0.650 |
| **RF** | 0.777 | 0.789 | 0.739 | 0.762 |
| **SVM** | 0.793 | 0.800 | 0.770 | 0.784 |
| **BERT** | 0.827 | 0.827 | 0.822 | 0.824 |

**Table 2.** Model performance with repeated holdout (average over five 70:30 train-test split).

| Model | Accuracy | Precision | Recall | F1-score |
|---|---|---|---|---|
| **LR** | 0.745 | 0.777 | 0.723 | 0.749 |
| **NB** | 0.766 | 0.800 | 0.750 | 0.770 |
| **DT** | 0.693 | 0.723 | 0.673 | 0.697 |
| **RF** | 0.758 | 0.802 | 0.718 | 0.758 |
| **SVM** | 0.773 | 0.818 | 0.733 | 0.773 |
| **BERT** | 0.844 | 0.836 | 0.845 | 0.840 |

As seen in Table 1 and Table 2, BERT outperforms traditional machine learning methods in classifying homophobic tweets across both cross-validation and holdout methodologies, demonstrating its nuanced context-capturing capability. LR showed improved performance in cross-validation compared to the holdout method, indicating better generalisation across different data subsets. In contrast, NB and DT fared better with the holdout method, suggesting potential overfitting in cross-validation scenarios; however, DT was the least effective overall, struggling with data variability. We observed that NB has higher accuracy and recall than BERT in the cross-validation experiments, but significantly lower precision and F1-score.

RF exhibited robust performance in both testing approaches, notably outshining others in cross-validation, which hints at its superior generalisability. SVM excelled among traditional models, particularly in cross-validation with an F1-score of 0.78, although performance dipped slightly in the holdout experiment, underscoring the benefits of diverse data exposure. Overall, BERT's superior performance in the holdout experiments underscores its effectiveness in handling extensive datasets, aligning with its advanced ability to understand context within text, marking it as a pivotal tool for detecting homophobic content.

## 6. Conclusion

Our work addresses a crucial gap in the current landscape of online hate speech detection by focusing on homophobia, an area that has remained underrepresented in sentiment analysis and general hate speech detection research. Through the development and evaluation of various machine learning models utilising BERT embeddings, including traditional machine learning methods like SVM and modern approaches like BERT, we have demonstrated the potential for nuanced detection of homophobic content on social media platforms such as X/Twitter.

Our approach has yielded promising results, with each of our models outperforming existing research results in this area by at least two-fold in both our cross-validation and repeated holdout experiments. This underscores the importance of collecting a rich representative dataset for domain-specific classification. The application of BERT embeddings, in particular, has shown advantages over traditional machine learning methods, highlighting the value of context-aware processing in identifying nuanced expressions of homophobia.

Looking forward, we acknowledge the need for continuous improvement and expansion of our research to encompass a broader spectrum of LGBTQIA+ hate speech. Despite the challenges faced, such as the difficulty in sourcing a comprehensive dataset, our research contributes to the broader effort of combatting online hate speech. The evolving nature of language and online discourse necessitates ongoing adaptation and refinement of detection models and strategies. It is also likely that exposing our model to a larger dataset could help it generalise better for this task. We hope that our research can contribute to creating a safer environment by providing a robust dataset that can be utilised by others to enhance their hate speech detection systems. Additionally, the availability of our data can stimulate further research in the field, leading to the development of more advanced and effective methods for identifying and mitigating hate speech online.

Although our research aims to produce resources to improve homophobia detection systems, it was also our intention to include transphobia detection within our study. However, we found it challenging to source sufficient and representative data. Therefore, we intend to extend our research to cover a wider range of LGBTQIA+ hate speech detection as part of future work in this area.

## 7. Acknowledgements

This work was conducted with the financial support of the Science Foundation Ireland Centre for Research Training in Artificial Intelligence under Grant No. 18/CRT/6223.

# 8. Appendices

*8.1. Appendix A*

We released our labelled dataset of tweets used to train our models for homophobia detection. Available at: https://huggingface.co/datasets/JoshMcGiff/HomophobiaDetectionTwitterX

*8.2. Appendix B*

We also released a version of our strongest BERT model from our holdout experiment. Available at: https://huggingface.co/JoshMcGiff/homophobiaBERT